\documentclass[11pt,a4paper]{article}
\usepackage[hyperref]{naaclhlt2019}
\usepackage{times}
\usepackage{latexsym}

\usepackage{url}
\usepackage{times}  
\usepackage{helvet}  
\usepackage{courier}  
\usepackage{url}  
\usepackage{graphicx}  

\usepackage{color}
\usepackage{mwe}
\usepackage{subfig}
\usepackage{amssymb}
\usepackage{mathtools}

\usepackage{latexsym}
\usepackage{multirow}
\usepackage[utf8]{inputenc}
\usepackage{dirtytalk}
\usepackage{multibib}
\usepackage{bm}
\usepackage{bbm}
\usepackage{amsmath}
\usepackage[linesnumbered,ruled]{algorithm2e}
\usepackage{amsfonts}
\usepackage{setspace}
\newcommand{\softmax}{\textit{softmax}}
\newcommand{\maximize}{\textit{maximize}}

\aclfinalcopy 

\title{Adapting RNN Sequence Prediction Model to Multi-label Set Prediction}

\author{Kechen Qin ~~~~ Cheng Li ~~~~ Virgil Pavlu ~~~~ Javed A. Aslam\\
  Khoury College of Computer Sciences \\
  Northeastern University\\
{\tt qin.ke@husky.neu.edu} ~~~~ {\tt \{chengli,vip,jaa\}@ccs.neu.edu}\\
   \\}

\date{}

\begin{document}
\maketitle
\begin{abstract}
We present an adaptation of RNN sequence models to the problem of multi-label classification for text, where the target is a \emph{set of labels}, not a sequence. Previous such RNN models define probabilities for sequences but not for sets; attempts to obtain a set probability are after-thoughts of the network design, including pre-specifying the label order, or relating the sequence probability to the set probability in \textit{ad hoc} ways.

Our formulation is derived from a principled notion of set probability, as the sum of probabilities of corresponding permutation sequences for the set. We provide a new training objective that maximizes this set probability, and a new prediction objective that finds the most probable set on a test document. These new objectives are theoretically appealing because they give the RNN model freedom to discover the best label order, which often is the natural one (but different among documents). 

We develop efficient procedures to tackle the computation difficulties involved in training and prediction. Experiments on benchmark datasets demonstrate that we outperform state-of-the-art methods for this task.

\end{abstract}

\section{Introduction}
\label{sec:Introduction}
Multi-label text classification is an important machine learning task wherein one must predict a set of labels to associate with a given document; for example, a news article might be tagged with labels \texttt{sport}, \texttt{football}, \texttt{2018 world cup}, and \texttt{Russia}. 
Formally, we are given a set
of label candidates $\mathcal{L}=\{1,2,...,L\}$, and we aim to build a classifier 
which maps a document $x$ to a set of labels $\mathbf{y}\subset \mathcal{L}$. The label set $\mathbf{y}$ is typically written as a binary vector $\mathbf{y}\in \{0,1\}^L$, with each bit $y_{\ell}$
indicating the presence or absence of a label.

Naively, one could predict each label independently without considering label dependencies. This approach is called Binary Relevance \cite{DBLP:journals/pr/BoutellLSB04,tsoumakas2007multi}, and is widely used due to its simplicity, but it often does not deliver good performance. Intuitively, knowing some labels---such as \texttt{sport} and \texttt{football}---should make it easier to predict \texttt{2018 world cup} and then \texttt{Russia}. There are several methods that try to capture label dependencies by building a joint probability estimation over all labels $p(\mathbf{y}=(y_1,y_2,...,y_L)|x)$ \cite{ghamrawi2005collective,read2009classifier,DBLP:conf/icml/DembczynskiCH10,li2016conditional}. 
The most popular approach, Probabilistic Classifier Chain (PCC) \cite{DBLP:conf/icml/DembczynskiCH10} learns labels one-by-one in a predefined fixed order: for each label, it uses one classifier to estimate the probability of that label given all previous labels predictions, $p(y_l|y_1,...,y_{l-1},x)$. PCC's well known drawback is that errors in early probability estimations tend to affect subsequent predictions, and can become massive when the total number of label candidates $L$ is large. 

Recurrent neural network (RNN) is originally designed to output a sequential structure, such as a sentence \cite{DBLP:conf/emnlp/ChoMGBBSB14}. Recently, RNNs have also been applied to multi-label classification by mapping the label set to a sequence \cite{DBLP:conf/cvpr/WangYMHHX16,DBLP:journals/corr/ZhangWSZL16,DBLP:conf/icpr/JinN16,DBLP:conf/iccv/WangCLXL17,DBLP:journals/corr/abs-1709-08553,DBLP:conf/aaai/ChenCYW18,DBLP:journals/corr/abs-1806-04822}. In contrast to PCC where a binary decision is made for each label sequentially, RNN only predicts the positive labels explicitly and therefore its decision chain length is equal to the number of positive labels, not the number of all labels. This makes RNN suffer less from early estimation errors than PCC.

Both PCC and RNN rely heavily on label orders in training and prediction. In multi-label data, the labels are given as sets, not necessarily with natural orders. RNN defines a sequence probability, while PCC defines set probability. Various ways of arranging sets as sequences have been explored: ordering alphabetically, by frequency, based on a label hierarchy, or according to some label ranking algorithm \cite{liu2015optimality}. Previous experimental results show that which order to choose can have a significant impact on learning and prediction \cite{vinyals2015order,DBLP:conf/nips/NamMKF17,DBLP:conf/aaai/ChenCYW18}. In the above example, starting label predictions sequence with \texttt{Russia}, while correct, would make the other predictions very difficult.

Previous work has shown that it is possible to train an RNN on multi-label data without specifying the label order in advance. With special training objectives, RNN can explore different label orders and converge to some order automatically \cite{vinyals2015order}. In this paper we follow the same line of study: We consider how to adapt RNN sequence model to multi-label set prediction without specifying the label order. Specifically, we make the following contributions:
\begin{enumerate}
\item We analyze existing RNN models proposed for multi-label prediction, and show that existing training and prediction objectives are not well justified mathematically and have undesired consequences in practice. 
\item We develop efficient approximate training and prediction methods. We propose new training and prediction objectives based on a principled notion of set probability. Our new formulation avoids the drawbacks of existing ones and gives the RNN model freedom to discover the best label order. 
\item We crawl two new datasets for multi-label prediction task, and apply our method to them. We also test our method on two existing multi-label datasets. The experimental results show that our method outperforms state-of-the-art methods on all datasets. We release the datasets at \url{http://www.ccis.neu.edu/home/kechenqin}. 
\end{enumerate}

\section{Mapping Sequences to Sets} 
In this section, we describe how existing approaches map sequences to sets, by writing down their objective functions using consistent notations. To review RNN designed for sequences, let $\mathbf{s}=(s_1,s_2,...,s_T)$ be an input sequence of outcomes, in a particular order, where $s_t \in \{1,2,...,L\}$; the order is often critical to the datapoint. An RNN model defines a probability distribution over all possible output sequences given the input in the form $p(\mathbf{s}=(s_1,s_2,...,s_T)|x)=\prod_{t=1}^T p(s_t|x,s_1,s_2,...,s_{t-1})$. To train the RNN model, one maximizes the likelihood of the ground truth sequence. 

At prediction time, one seeks to find the sequence with the highest probability $\mathbf{s}^*=\arg\max_\mathbf{s} p(\mathbf{s}|x)$, and this is usually implemented approximately with a beam search procedure \cite{lowerre1976harpy} (we modified into Algorithm \ref{alg:beam_combine}). The sequence history is encoded with an internal memory vector $h_t$ which is updated over time. RNN is also often equipped with the attention mechanism \cite{DBLP:journals/corr/BahdanauCB14}, which in each timestep $t$ puts different weights on different words (features) and thus effectively attends on a list of important words. The context vector $c_t$ is computed as the weighted average over the dense representation of important words to capture information from the document. The context $c_t$, the RNN memory $h_t$ at timestep $t$, and the encoding of previous label ${s_{t-1}}$ are all concatenated and used to model the label probability distribution at time $t$ as $p(s_t|x,s_1,s_2,...,s_{t-1}) \sim \softmax(\phi(c_t,h_t,s_{t-1}))$, where $\phi$ is a non-linear function, and $\softmax$ is the normalized exponential function. 

 \begin{table*}[t]
 	\begin{center}

\resizebox{1.0\textwidth}{!}{%
\hspace{-2ex}
 \begin{tabular}{|c|c|c|}
 \hline
 Methods	&Training objectives	& Prediction objectives\\
 \hline
 seq2seq-RNN & $\maximize \sum_{n=1}^N \log p(\mathbf{s}^{(n)}|x^{(n)})$ & $\hat{\mathbf{y}}=set(\mathbf{s}^*)$, $\mathbf{s}^*=\arg\max_{\mathbf{s}} p(\mathbf{s}|x)$\\
 \hline 
 Vinyals-RNN-max & $\maximize \sum_{n=1}^N \max_{\mathbf{s}\in \pi(\mathbf{y}^{(n)})}\log p(\mathbf{s}|x^{(n)})$ & $\hat{\mathbf{y}}=set(\mathbf{s}^*)$, $\mathbf{s}^*=\arg\max_{\mathbf{s}} p(\mathbf{s}|x)$\\
 \hline
 Vinyals-RNN-uniform & $\maximize \sum_{n=1}^N \sum_{\mathbf{s}\in \pi(\mathbf{y}^{(n)})}\log p(\mathbf{s}|x^{(n)})$ & $\hat{\mathbf{y}}=set(\mathbf{s}^*)$, $\mathbf{s}^*=\arg\max_{\mathbf{s}} p(\mathbf{s}|x)$\\
 \hline
 Vinyals-RNN-sample & $\maximize \sum_{n=1}^N \sum_{\mathbf{s}\in \pi(\mathbf{y}^{(n)})}p(\mathbf{s}|x^{(n)})\log p(\mathbf{s}|x^{(n)})$ & $\hat{\mathbf{y}}=set(\mathbf{s}^*)$, $\mathbf{s}^*=\arg\max_{\mathbf{s}} p(\mathbf{s}|x)$\\
 \hline
 set-RNN (ours) & $\maximize \sum_{n=1}^N \log \sum_{\mathbf{s}\in \pi(\mathbf{y}^{(n)})} p(\mathbf{s}|x^{(n)})$& $\hat{\mathbf{y}}=\arg\max_{\mathbf{y}} p(\mathbf{y}|x)$\\
 \hline
 \end{tabular}
 }
 \caption{Comparison between previous and our \emph{set-RNN} training and prediction objectives.} \label{tab_objs}
 \end{center}
 \end{table*}

 To apply RNN to multi-label problems, one approach is to map the given set of labels $\mathbf{y}$ to a sequence $\mathbf{s}=(s_1,s_2,...,s_T)$, on training documents. This is usually obtained by writing the label set in a globally fixed order (e.g.\ by label frequency), as in PCC.
 Once the mapping is done, RNN is trained with the standard maximum likelihood objective \cite{DBLP:conf/nips/NamMKF17}: 
\begin{align}
\maximize \sum_{n=1}^N \log p(\mathbf{s}^{(n)}|x^{(n)})
\label{eq:standard_rnn}
\end{align}
where $x^{(n)}$ is the $n$-th document and $N$ is the total number of documents in the corpus. 

\newcite{vinyals2015order} proposes to dynamically choose during training the sequence order deemed as most probable by the current RNN model:
\begin{align}
\maximize \sum_{n=1}^N \max_{\mathbf{s}\in \pi(\mathbf{y}^{(n)})}\log p(\mathbf{s}|x^{(n)})
\label{eq:max_obj}
\end{align}
where the $\pi(\mathbf{y}^{(n)})$ stands for all permutations of the label set $\mathbf{y}^{(n)}$. This eliminates the need to manually specify the label order.
However, as noticed by the authors, this objective cannot be used in the early training stages: the early order choice (often random) is reinforced by this objective and can be stuck upon permanently. To address this issue, \newcite{vinyals2015order}~also proposes two smoother alternative objectives to initialize the model training:

The authors suggest that one first consider many random orders for each label set in order to explore the space:
\begin{align}
\maximize \sum_{n=1}^N \sum_{\mathbf{s}\in \pi(\mathbf{y}^{(n)})}\log p(\mathbf{s}|x^{(n)})
\label{eq:wrong_obj}
\end{align} 

After that, one can sample sequences following the model predictive distribution instead of uniform distribution:
\begin{align}
\maximize \sum_{n=1}^N \sum_{\mathbf{s}\in \pi(\mathbf{y}^{(n)})}p(\mathbf{s}|x^{(n)})\log p(\mathbf{s}|x^{(n)})
\label{eq:sample_obj}
\end{align} 

In training, one needs to schedule the transition among these objectives, a rather tricky endeavor. At prediction time, one needs to find the most probable set. This is done by (approximately) finding the most probable sequence $\mathbf{s}^*=\arg\max_\mathbf{s} p(\mathbf{s}|x)$ and treating it as a set $\hat{\mathbf{y}}=set(\mathbf{s}^*)$. With a large number of sequences, it is quite possible that the argmax has actually a low probability, which can lead to neglecting important information when we ignore sequences other than the top one.





\section{Adapting RNN Sequence Prediction Model to Multi-label Set Prediction}
\label{sec:Model}

We propose a new way of adapting RNN to multi-label set prediction, which we call \emph{set-RNN}. We appreciate the RNN model structure \cite{rumelhart1988learning} (defines a probability distribution over all possible sequences directly) and introduce training and prediction objectives tailored for sets that take advantage of it, while making a clear distinction between the sequence probability $p(\mathbf{s}|x)$ and the set probability $p(\mathbf{y}|x)$. 
 We define the set probability as the sum of sequences probabilities for all sequence permutations of the set, namely $p(\mathbf{y}|x)=\sum_{\mathbf{s}\in \pi(\mathbf{y})} p(\mathbf{s}|x)$. Based on this formulation, an RNN also defines a probability distribution over all possible sets indirectly since $\sum_{\mathbf{y}} p(\mathbf{y}|x)=\sum_{\mathbf{y}}\sum_{\mathbf{s}\in \pi(\mathbf{y})} p(\mathbf{s}|x)=\sum_{\mathbf{s}} p(\mathbf{s}|x)=1$.  (For this equation to hold, in theory, we should also consider permutations $\mathbf{s}$ with repeated labels, such as $(1,2,3,1)$. But in practice, we find it very rare for RNN to actually generate sequences with repeated labels in our setup, and whether allowing repetition or not does not make much difference.)

 In standard maximum likelihood training, one wishes to maximize the likelihood of given label sets, namely, $\prod_{n=1}^N p(\mathbf{y}^{(n)}|x^{(n)})=\prod_{n=1}^N \sum_{\mathbf{s}\in \pi(\mathbf{y}^{(n)})} p(\mathbf{s}|x^{(n)})$, or equivalently, 
\begin{align}
\maximize \sum_{n=1}^N \log \sum_{\mathbf{s}\in \pi(\mathbf{y}^{(n)})} p(\mathbf{s}|x^{(n)})
\label{eq:right_obj}
\end{align}



\subsection{How is our new formulation different?}

This training objective (\ref{eq:right_obj}) looks similar to the objective (\ref{eq:wrong_obj}) considered in previous work \cite{vinyals2015order}, but in fact they correspond to very different transformations. Under the maximum likelihood framework, our objective (\ref{eq:right_obj}) corresponds to the transformation $p(\mathbf{y}|x)=\sum_{\mathbf{s}\in \pi(\mathbf{y})} p(\mathbf{s}|x)$, while objective (\ref{eq:wrong_obj}) corresponds to the transformation $p(\mathbf{y}|x)=\prod_{\mathbf{s}\in \pi(\mathbf{y})} p(\mathbf{s}|x)$. The latter transformation does not define a valid probability distribution over $\mathbf{y}$ (i.e., $\sum_{\mathbf{y}} p(\mathbf{y}|x)\neq 1$), and it has an undesired consequence in practical model training: because of the multiplication operation, the RNN model has to assign equally high probabilities to all sequence permutations of the given label set in order to maximize the set probability. If only some sequence permutations receive high probabilities while others receive low probabilities, the set probability computed as the product of sequence probabilities will still be low. In other words, if for each document, RNN finds one good way of ordering relevant labels (such as hierarchically) and allocates most of the probability mass to the sequence in that order, the model still assigns low probabilities to the ground truth label sets and will be penalized heavily. As a consequence the model has little freedom in discovering and concentrating on some natural label order. In contrast, with our proposed training objective, in which the multiplication operation is replaced by the summation operation, it suffices to find only one reasonable permutation of the labels for each document. It is worth noting that different documents can have different label orders; thus our proposed training objective gives the RNN model far more freedom on label order. The other two objectives (\ref{eq:max_obj}) and (\ref{eq:sample_obj}) proposed in \cite{vinyals2015order} are less restrictive than (\ref{eq:wrong_obj}), but they have to work in conjunction with (\ref{eq:wrong_obj}) because of the self reinforcement issue. Our proposed training objective has a natural probabilistic interpretation, and does not suffer from self reinforcement issue. Thus it can serve as a stand alone training objective. Also, using Jensen’s inequality, one can show that objective (\ref{eq:wrong_obj}) is maximizing a lower bound on the log-likelihood, while objective (\ref{eq:right_obj}) is maximizing it directly. 

\begin{algorithm}[ht]
  \SetKwInOut{Input}{Input}
  \SetKwInOut{Output}{Output}

  \Input{Instance $x$ \\
  Subset of labels considered $G\subset \mathcal{L}$\\
  Boolean flag $ALL$: 1 if sequences must contain all $G$ labels; 0 if partial sequences are allowed }
  \Output{A list of top sequences and the associated probabilities} 
  Let $\mathbf{s}_1$,$\mathbf{s}_2$,...,$\mathbf{s}_K$ be the top $K$ sequences found so far. Initially, all $K$ sequences are empty. $\oplus$ means concatenation. \\
  \While {true}{
   // Step 1: Generate Candidate Sequences from each existing sequence $s_k \in K$ and all possible new labels $l \in G$:\\
  Expand all non-stopped sequences:\\
   $C = \{ \mathbf{s}_k\oplus l | l\in G, STOP \notin s_k \}$\\  
Include stopped sequences:\\
  $C = C \cup \{ \mathbf{s}_k | STOP \in s_k \}$\\
  Terminate non-stopped sequences:\\
  \If{$ALL==0$}{
  %
  $C =C \cup \{ \mathbf{s}_k\oplus STOP | STOP \notin s_k \}$
  }


 // Step 2: Select highest probabilities sequences from candidate set $C$\\
  $K$
  = topK-argmax$_k \{\text{prob}[s_k] | s_k \in C\}$\\
  \If {all top $K$ sequences end with $STOP$ or contain all labels in $G$}{
  Terminate the algorithm}

   }
  \Return{sequence list $\mathbf{s}_1$,$\mathbf{s}_2$,...,$\mathbf{s}_K$ and the associated probabilities}
  \caption{Beam\_Search}
  \label{alg:beam_combine}
\end{algorithm}

\subsection{Training by Maximizing Set Probability}
Training an RNN model with the proposed objective (\ref{eq:right_obj}) requires summing up sequence (permutation) probabilities for a set $\mathbf{y}$, where $|\mathbf{y}|$ is the cardinality of the set. Thus evaluating this objective exactly can be intractable. We can approximate this sum by only considering the top $K$ highest probability sequences produced by the RNN model. We introduce a variant of beam search for sets with width $K$ and with the search candidates in each step restricted to only labels in the set (see Algorithm~\ref{alg:beam_combine} with $ALL=1$)
. This approximate inference procedure is carried out repeatedly before each batch training step, in order to find highest probability sequences for all training instances occurring in that batch. The overall training procedure is summarized in Algorithm \ref{alg:train_beam}.

\begin{algorithm}
  \SetKwInOut{Input}{Input}
  \SetKwInOut{Output}{Output}

  \Input{Multi-label dataset $(x^{(n)},\mathbf{y}^{(n)}),n=1,2,...,N$ }
  \Output{Trained RNN model parameters}
  
  \ForEach {batch}{
      \ForEach{$(x^{n},\mathbf{y}^{n})$ in the batch}{
        Get top $K$ sequences :\\
		  	$\{\mathbf{s}^n_{1},...,\mathbf{s}^n_{K}, p(\mathbf{s}^n_{1}|x^n),...,p(\mathbf{s}^n_{K}|x^n)\}$= \hspace{4ex}= Beam\_Search$(x^{n},\mathbf{y}^{n}, ALL=1$)\\
        }
      Update model parameters by maximizing $\sum\limits_{(x^{n},\mathbf{y}^{n}) \in \text{batch}} \log \sum\limits_{\mathbf{s}\in\{\mathbf{s}^n_{1},...,\mathbf{s}^n_{K}\}} p(\mathbf{s}|x^{n})$
  }
  \caption{Training method for set-RNN}
  \label{alg:train_beam}
\end{algorithm}

\subsection{Predicting the Most Probable Set}
The transformation $p(\mathbf{y}|x)=\sum_{\mathbf{s}\in \pi(\mathbf{y})} p(\mathbf{s}|x)$ also naturally leads to a prediction procedure, which is different from the previous standard of directly using most probable sequence as a set. We instead aim to find the most likely set $\hat{\mathbf{y}}=\arg\max_{\mathbf{y}} p(\mathbf{y}|x)$, which involves summing up probabilities for all of its permutations. To make it tractable, we propose a two-level beam search procedure. First we run standard RNN beam search (Algorithm \ref{alg:beam_combine} with $ALL=0$) to generate a list of highest probability sequences. We then consider the label set associated with each label sequence. For each set, we evaluate its probability using the same approximate summation procedure as the one used during model training (Algorithm~\ref{alg:beam_combine} with $ALL=1$): we run our modified beam search to find the top few highest probability sequences associated with the set and sum up their probabilities. Among these sets that we have evaluated, we choose the one with the highest probability as the prediction. The overall prediction procedure is summarized in Algorithm~\ref{alg:test}. As we shall show in case study, the most probable set may not correspond to the most probable sequence; these are certainly cases where our method has an advantage.

Both our method and the competitor state-of-the-art (Vinyals-RNNs) are at most $K$ times slower than a vanilla-RNN,  due to the time spent on dealing with $K$ permutations per datapoint. Our proposed method is about as fast as the Vinyals-RNN methods, except for the Vinyals-RNN-uniform which is a bit faster (by a factor of 1.5) because its epochs do not run the additional forward pass. 

%
%

\begin{algorithm}
  \SetKwInOut{Input}{Input}
  \SetKwInOut{Output}{Output}

  \Input{Instance $x$}
  \Output{Predicted label set $\hat{\mathbf{y}}$}
  Obtain $K$ highest probability sequences :\\
  $\{\mathbf{s}_1,...,\mathbf{s}_{K}\}$ =  Beam\_Search(x,$\mathcal{L}, ALL=0$)\\
  
  Map each sequence $\mathbf{s}_k$ to the corresponding set $\mathbf{y}_k$ and remove duplicate sets (if any)
  
  \ForEach {$\mathbf{y}_k$}{
    
     Get $K$ most probable sequences associated with $\mathbf{y}_k$ and their probabilities :\\
	  $\{\mathbf{s'}_{1},...,\mathbf{s'}_{K}, p(\mathbf{s}'_{1}|x),...,p(\mathbf{s}'_{K}|x)\}$= \\ \hspace{6ex} = Beam\_Search(x,$\mathbf{y}_k, ALL=1$)\\
    Set probability is approx by summing up :
    $p({\mathbf{y}_k}|x) \approx \sum\limits_{\mathbf{s}\in \{\mathbf{s}'_{1},...,\mathbf{s}'_K\}} p(\mathbf{s}|x)$
  }  
  $\hat{\mathbf{y}} = argmax_{{\mathbf{y}_k}}(p({\mathbf{y}_k}|x))$
  \caption{Prediction Method for set-RNN}
  \label{alg:test}
\end{algorithm}



 
 
 


\section{Results and Analysis}
\label{sec:results}
\subsection{Experimental Setup}

We test our proposed set-RNN method on 4 real-world datasets, RCV1-v2, Slashdot, TheGuardian, and Arxiv Academic Paper Dataset (AAPD) \cite{DBLP:journals/corr/abs-1806-04822}. We take the public RCV1-v2 release\footnote{\scriptsize\url{http://www.ai.mit.edu/projects/jmlr/papers/volume5/lewis04a/lyrl2004_rcv1v2_README.htm}} and randomly sample 50,000 documents. We crawl Slashdot and TheGuardian documents from their websites\footnote{\scriptsize Slashdot: \url{https://slashdot.org/} Note that there is another public Slashdot multi-label dataset \cite{read2009classifier} but we do not use that one because it is quite small. TheGuardian: \url{https://www.theguardian.com} 
} and treat the official editor tags as ground truth. We also gather a list of user tags\footnote{\scriptsize \url{www.zubiaga.org/datasets/socialbm0311/}} for each document and treat them as additional features. For AAPD dataset, we follow the same train/test split as in \cite{DBLP:journals/corr/abs-1806-04822}. Table \ref{tab:stats} contains statistics of these four datasets. Links to document, official editor tags, and user tags are avaliable at \url{http://www.ccis.neu.edu/home/kechenqin}. 
\begin{table}[h]
	\resizebox{1.01\columnwidth}{!}{
\begin{tabular}{|l|c|c|c|c|c|c|}
 \hline
 Data & \#Train & \#Test & Cardinality & \#Labels & Doc length \\
 \hline
 Slashdot & 19,258 & 4,814& 4.15& 291&64 \\
 RCV1-v2 & 40,000& 10,000& 3.17& 101&121 \\
 TheGuardian & 37,638 & 9,409 & 7.41& 1,527&505 \\
 AAPD & 53,840 & 1,000 & 2.41& 54 & 163 \\
 \hline
\end{tabular}
}
\caption{\fontsize{10}{12}\selectfont Statistics of the datasets.}\label{tab:stats}
\end{table}

\begin{table*}[ht]
	\begin{center}
	\resizebox{2.1\columnwidth}{!}{
		\begin{tabular}{|l|cc|cc|cc|cccc|}
			\hline
			\multirow{2}{*}{Methods}	& 
       \multicolumn{2}{c|}{Slashdot}&\multicolumn{2}{c|}{RCV1-v2}&\multicolumn{2}{c|}{TheGuardian}&\multicolumn{4}{c|}{AAPD} \\
       \cline{2-11}
      & label-F1 & instance-F1 & label-F1 & instance-F1 & label-F1 & instance-F1 & label-F1 & instance-F1 & hamming-loss & micro-F1 \\

            \hline
      BR
      & .271 & .484 & .486 & .802 &.292 & .572 & .529 & .654 & .0230 & .685 \\
      BR-support
      & .247 & .516 & .486 & .805 &.296 & .594 & .545 & .689 & \textbf{.0228} & .696\\
      PCC 
      & .279 & .480 & .595 & .818 &- & - & .541 & .688 & .0255 & .682 \\
      seq2seq-RNN 
       & .270 & .528 & .561 & .824 &.331 & .603 & .510 & .708 & .0254 & .701 \\
      Vinyals-RNN-uniform 
       & .279 & .527 & .578 & .826 & .313 & .567 & .532 & .721 & .0241 & .711\\
      Vinyals-RNN-sample
      & .300 & .531 & .590 & .828 & .339 & .597 & .527 & .706 & .0259 & .697 \\
      Vinyals-RNN-max
      & .293 & .530 & .588 & .829 & .343 & .599 & .535 & .709 & .0256 & .700 \\
      Vinyals-RNN-max-direct
       & .226 & .518 & .539 & .808 & .313 & .583 & .490 & .702 & .0257 & .694\\
      SGM
      &-&-&-&-&-&-& - & - & .0245 & .710 \\
      set-RNN 
       & \textbf{.310} & \textbf{.538} & \textbf{.607} & \textbf{.838} &\textbf{.361} & \textbf{.607} & \textbf{.548} & \textbf{.731} & .0241 & \textbf{.720}\\
      \hline

		\end{tabular}
		}
	\end{center}
	\caption{Comparison of different approaches. ``-'' means result not available. For \emph{hamming loss}, the lower the value is, the better the model performs. For all other measures, the higher the better.
  }\label{tab:main}
\end{table*}

\begin{table*}[ht]
 \begin{center}
 	\resizebox{2.1\columnwidth}{!}{

  \begin{tabular}{|l|cc|cc|cc|cc|}
   \hline
   \multirow{2}{*}{Methods} & 
       \multicolumn{2}{c|}{Slashdot}&\multicolumn{2}{c|}{RCV1-v2}&\multicolumn{2}{c|}{TheGuardian}&\multicolumn{2}{c|}{AAPD} \\
       \cline{2-9}
      & label-F1 & instance-F1 & label-F1 & instance-F1 & label-F1 & instance-F1 & label-F1 & instance-F1\\

            \hline
      seq2seq-RNN 
       & .270$\to$.269 & .528$\to$.528& .561$\to$.561 & .824$\to$.824 &.331$\to$.336& .603$\to$.603 & .510$\to$.511 &.708$\to$.709 \\
      Vinyals-RNN-uniform 
       & .279$\to$.288& \textbf{.527}$\to$\textbf{.537}& .578$\to$.587& .826$\to$.833& \textbf{.313}$\to$\textbf{.336}& \textbf{.567}$\to$\textbf{.585} & \textbf{.532}$\to$\textbf{.542} &.721$\to$.724 \\
       Vinyals-RNN-sample
      & .300$\to$.303 & .531$\to$.537 &.590$\to$.597 & .828$\to$.833 & \textbf{.339}$\to$\textbf{.351} & .597$\to$.602 & .527$\to$.530 & .706$\to$.708 \\
      Vinyals-RNN-max
      & .293$\to$.301 & .530$\to$.535 & .588$\to$.585 & .829$\to$.830 & .343$\to$.352 &.599$\to$.604 & .535$\to$.537 &.709$\to$.712 \\
      Vinyals-RNN-max-direct
       & .226$\to$.228& .518$\to$.519& .539$\to$.538& .808$\to$.808& .313$\to$.316&.583$\to$.584 & .490$\to$.490 &.702$\to$.701 \\
      
      set-RNN 
       & \textbf{.297}$\to$\textbf{.310} & \textbf{.528}$\to$\textbf{.538} & \textbf{.593}$\to$\textbf{.607} & .831$\to$.838 &\textbf{.349}$\to$\textbf{.361} & \textbf{.595}$\to$\textbf{.607} & .548$\to$.548 &.728$\to$.731 \\\hline

  \end{tabular}
  }
 \end{center}
 \caption{Predicting the most probable sequence vs. predicting the most probable set. Numbers before the arrow: predicting the most probable sequence. Numbers after the arrow: predicting the most probable set. We highlight scores which get significantly improved in \textbf{bold} (improvement is larger than 0.01).
  }\label{table_prediction}
\end{table*}

 To process documents, we filter out stopwords and punctuations. Each document is truncated to have maximum 500 words for TheGuardian and AAPD, and 120 for Slashdot and RCV1-v2. Zero padding is used if the document contains less words than the maximum number. Numbers and out-of-vocabulary words are replaced with special tokens. Words, user tags and labels are all encoded as 300-dimensional vectors using \textsc{word2vec} \cite{DBLP:journals/corr/abs-1301-3781}.
 
 We implement RNNs with attention using \textsc{tensorflow-1.4.0} \cite{DBLP:conf/osdi/AbadiBCCDDDGIIK16}. The dynamic function for RNNs is chosen to be Gated recurrent units (GRU) with 2 layers and at most 50 units in decoder. The size of the GRU unit is 300. We set dropout rate to 0.3, and train the model with Adam optimizer \cite{DBLP:journals/corr/KingmaB14} with learning rate $0.0005$. Beam size is set to be 12 at both training and inference stages. We adopt \emph{label-F1} (average F1 over labels) and \emph{instance-F1} (average F1 over instances) as our main evaluation metrics, as defined below:
 
\begin{align*} \text{label-F1} = \frac{1}{L}\sum_{\ell=1}^L\frac{2\sum_{n=1}^N y^{(n)}_\ell \hat{y}^{(n)}_\ell}{\sum_{n=1}^N y^{(n)}_\ell+\sum_{n=1}^N \hat{y}^{(n)}_\ell}\\
\text{instance-F1} = \frac{1}{N}\sum_{n=1}^N\frac{2\sum_{\ell=1}^L y^{(n)}_\ell \hat{y}^{(n)}_\ell}{\sum_{\ell=1}^L y^{(n)}_\ell+\sum_{\ell=1}^L \hat{y}^{(n)}_\ell}
\end{align*}
where for each instance $n$, $y_\ell^{(n)}=1$ if label $\ell$ is a given label in ground truth; $\hat{y}_\ell^{(n)}=1$ if label $\ell$ is a predicted label.

We compare our method with the following methods: 
\begin{itemize}
	\item \textbf{Binary Relevance (BR)} \cite{tsoumakas2007multi} with both independent training and prediction;
	\item \textbf{Binary Relevance with support inference (BR-support)} \cite{wang2018pipeline} which trains binary classifiers independently but imposes label constraints at prediction time by only considering label sets observed during training, namely $\hat{\mathbf{y}}=\arg\max_{\text{observed~}\mathbf{y}}\prod_{\ell=1}^L p(y_{\ell}|x)$;
	\item \textbf{Probabilistic Classifier Chain (PCC)} \cite{DBLP:conf/icml/DembczynskiCH10} which transforms the multi-label classification task into a chain of binary classification problems. Predictions are made with Beam Search.
  \item \textbf{Sequence to Sequence RNN (seq2seq-RNN)} \cite{DBLP:conf/nips/NamMKF17} which maps each set to a sequence by decreasing label frequency and solves the multi-label task with an RNN designed for sequence prediction (see Table \ref{tab_objs}). 
  \item \textbf{Vinyals-RNN-uniform, Vinyals-RNN-sample, and Vinyals-RNN-max} are three variants of RNNs proposed by \cite{vinyals2015order}. They are trained with different objectives that correspond to different transformations between sets and sequences. See Table~\ref{tab_objs} for a summary of their training objectives. Following the approach taken by \cite{vinyals2015order}, Vinyals-RNN-sample and Vinyals-RNN-max are initialized by Vinyals-RNN-uniform. We have also tested training Vinyals-RNN-max directly without having Vinyals-RNN-uniform as an initialization, and we name it as \textbf{Vinyals-RNN-max-direct}.
  \item \textbf{Sequence Generation Model (SGM)} \cite{DBLP:journals/corr/abs-1806-04822} which trains the RNN model similar to seq2seq-RNN but uses a new decoder structure that computes a weighted global embedding based on all labels as opposed to just the top one at each timestep.
\end{itemize}

In BR and PCC, logistic regressions with L1 and L2 regularizations are used as the underlying binary classifiers. seq2seq-RNN, PCC, and SGM rely on a particular label order. We adopt the decreasing label frequency order, which is the most popular choice.




\subsection{Experimental Results}


Table \ref{tab:main} shows the performance of different methods in terms of \emph{label-F1} and \emph{instance-F1}. The SGM results are taken directly from \cite{DBLP:journals/corr/abs-1806-04822}, and are originally reported only on AAPD dataset in terms of \emph{hamming-loss} and \emph{micro-F1}. Definitions of these two metrics can be found in \cite{koyejo2015consistent}. 

Our method performs the best in all metrics on all datasets (except hamming loss on AAPD, see table \ref{tab:main}). In general, RNN based methods perform better than traditional methods BR, BR-support and PCC. Among the Vinyals-RNN variants, Vinyals-RNN-max and Vinyals-sample work the best and have similar performance. However, they have to be initialized by Vinyals-RNN-uniform. Otherwise, the training gets stuck in early stage and the performance degrades significantly. One can see the clear degradation by comparing the Vinyals-RNN-max row (with initialization) with the Vinyals-RNN-max-direct row (without initialization). By contrast, our training objective in set-RNN does not suffer from this issue and can serve as a stable stand alone training objective.

\begin{figure}[t]
\includegraphics[width=1.0\columnwidth]{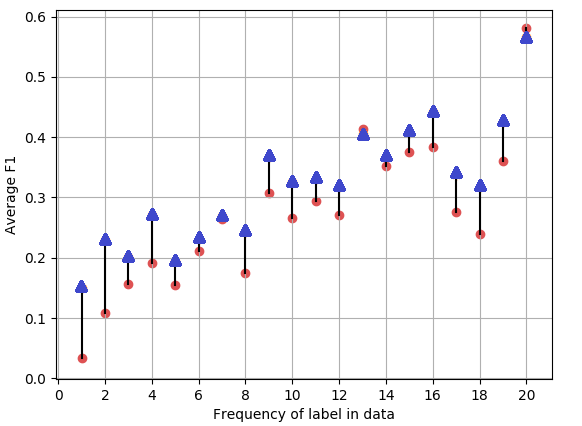}

\caption{Average F1 over rare labels with the same frequency on TheGuardian dataset. Blue($\Delta$)=set-RNN, Red($\cdot$)=seq2seq-RNN.}
\label{fig:labelf1}
\end{figure}

On TheGuardian dataset, set-RNN performs slightly better than seq2seq-RNN in terms of instance-F1, but much better in terms of label-F1. It is known that instance-F1 is basically determined by the popular labels' performance while label-F1 is also sensitive to the performance on rare labels. Figure~\ref{fig:labelf1} shows that set-RNN predicts rare labels better than seq2seq-RNN.

Next we analyze how much benefit our new set prediction strategy brings in. For each RNN-based method, we test two prediction strategies: 1) finding the sequence with the highest probability and outputting the corresponding set (this is the default prediction strategy for all models except set-RNN); 2) outputting the set with the highest probability (this is the default prediction strategy for set-RNN). Table~\ref{table_prediction} shows how each method performs with these two prediction strategies. One can see that Vinyals-RNN-uniform and set-RNN benefit most from predicting the top set, Vinyals-RNN-sample, Vinyals-RNN-max and Vinyals-RNN-max-direct benefit less, and seq2seq RNN does not benefit at all. Intuitively, for the top-set prediction to be different from the top-sequence prediction, the model has to spread probability mass across different sequence permutations of the same set. 

\subsection{Analysis: Sequence Probability Dsitribution}
\begin{figure}[t]
\includegraphics[width=1.00\columnwidth]{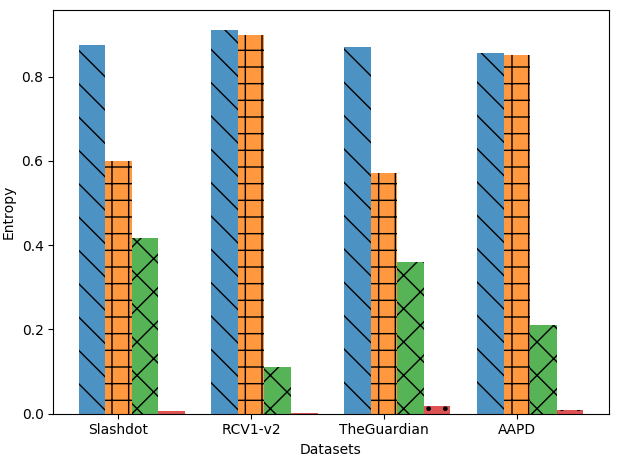}

\caption{Entropy of sequence probability distribution for each model. Blue(\textbackslash)=Vinyals-RNN-uniform, Orange(+)=set-RNN, Green($\times$)=Vinyals-RNN-max, Red($\cdot$)=seq2seq-RNN.}
\label{fig:entropy}
\end{figure}

Results in Table~\ref{table_prediction} motivates us to check how sharply (or uniformly) distributed the probabilities are over different sequence permutations of the predicted set. We first normalize these sequence probabilities related to the predicted set and then compute the entropy. To make predictions with different set sizes (and hence different number of sequence permutations) comparable, we further divide the entropy by the logarithm of number of sequences. Smaller entropy values indicate a sharper distributions. The results are shown in Figure~\ref{fig:entropy}.

seq2seq-RNN trained with fixed label order and standard RNN objective (\ref{eq:standard_rnn}) generates very sharp sequence distributions. It basically only assigns probability to one sequence in the given order. The entropy is close to 0. In this case, predicting the set is no different than predicting the top sequence (see Table~\ref{table_prediction}). On the other extreme is Vinyals-RNN-uniform, trained with objective (\ref{eq:wrong_obj}), which spreads probabilities across many sequences, and leads to the highest entropy among all models tested (the uniform distribution has the max entropy of 1). From Table~\ref{table_prediction}, we see that by summing up sequence probabilities and predicting the most probable set, Vinyals-RNN-uniform's performance improves. But as discussed earlier, training with the objective (\ref{eq:wrong_obj}) makes it impossible for the model to discover and concentrate on a particular natural label order (represented by a sequence). Overall Vinyals-RNN-uniform is not competitive even with the set-prediction enhancement. Between the above two extremes are Vinyals-RNN-max and set-RNN (we have omitted Vinyals-RNN-sample and Vinyals-RNN-max-direct here as they are similar to Vinyals-RNN-max). Both models are allowed to assign probability mass to a subset of sequences. Vinyals-RNN-max produces sharper sequence distributions than set-RNN, because Vinyals-RNN-max has the incentive to allocate most of the probability mass to the most probable sequence due to the max operator in its training objective (\ref{eq:max_obj}). From Table~\ref{table_prediction}, one can see that set-RNN clearly benefits from summing up sequence probabilities and predicting the most probable set while Vinyals-RNN-max does not benefit much. Therefore, the sequence probability summation is best used in both training and prediction, as in our proposed method.

Comparing 4 datasets in Table~\ref{table_prediction}, we also see that Slashdot and TheGuardian, which have larger label cardinalities (therefore more permutations for one set potentially), benefit more from predicting the most probable set than RCV1 and AAPD, which have smaller label cardinalities.

\section{Case Analysis}
We further demonstrate how set-RNN works with two examples.
In the first example from the RCV1-v2 dataset, the most probable set predicted by set-RNN (which is also the correct set in this example) does not come from the most probable sequence. Top sequences in decreasing probability order are listed in Table~\ref{tab:pred_case_study}. The correct label set \{forex, markets, equity, money markets, metals trading, commodity\} has the maximum total probability of 0.161, but does not match the top sequence.

\begin{table}[h]
\resizebox{1.01\columnwidth}{!}{
\begin{tabular}{l|l}
PROB	& SEQUENCE\\
\hline
0.0236 	& equity, markets, money markets, forex\\
0.0196	& {\bf forex, markets, equity, money markets, metals trading, commodity}\\
0.0194	& {\bf equity, markets, forex, money markets, metals trading, commodity}\\
0.0159	& {\bf markets, equity, forex, money markets, metals trading, commodity}\\
0.0157	& {\bf forex, money markets, equity, metals trading, markets, commodity}\\
0.0153	& {\bf forex, money markets, markets, equity, metals trading, commodity}\\
0.0148	& markets, equity, money markets, forex\\
0.0143	& {\bf money markets, equity, metals trading, commodity, forex, markets}\\
0.0123	& {\bf markets, money markets, equity, metals trading, commodity, forex}\\
0.0110 	& {\bf markets, equity, forex, money markets, commodity, metals trading}\\
0.0107	& {\bf forex, markets, equity, money markets, commodity, metals trading}\\
0.0094	& {\bf forex, money markets, equity, markets, metals trading, commodity}\\
\hline
\end{tabular}
}
\caption{The set-RNN predicted set (also the correct set) \{forex, markets, equity, money markets, metals trading, commodity\} has the max total probability of 0.161, but does not match the top sequence. Sequences for the correct set are in bold.}\label{tab:pred_case_study}
\end{table}

\begin{figure}[t]
\includegraphics[width=1.0\columnwidth]{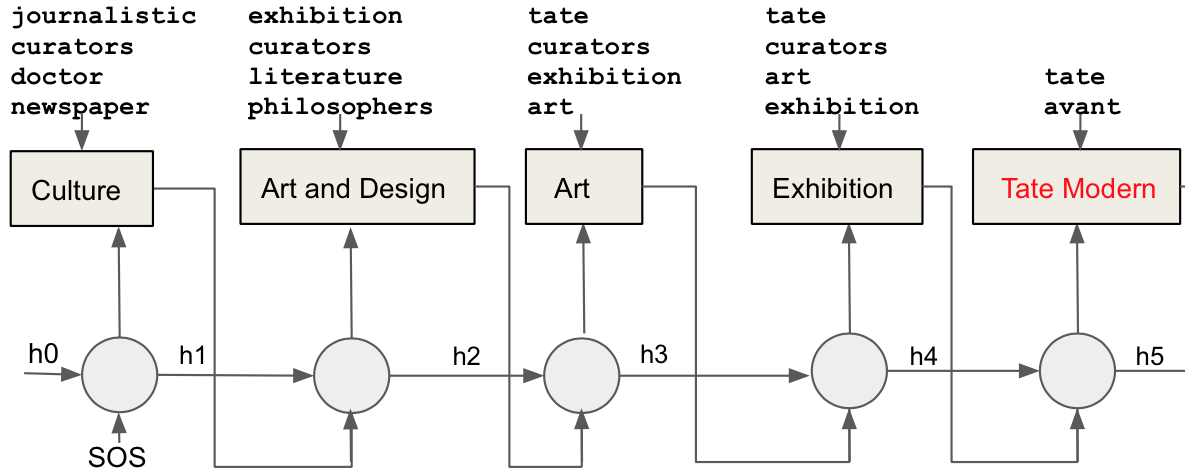}
\includegraphics[width=1.0\columnwidth]{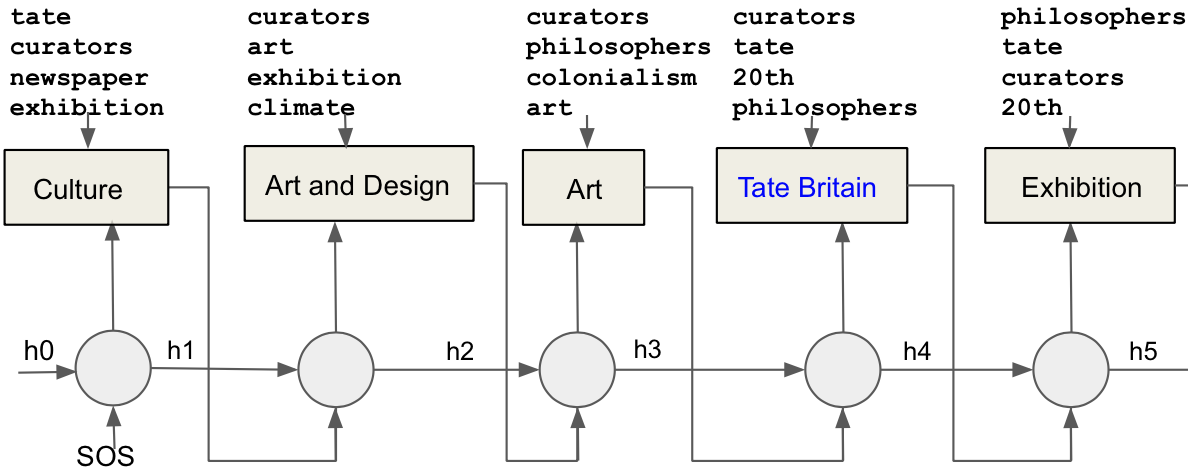}

\caption{\fontsize{10}{12}\selectfont Top: best sequence by seq2seq-RNN; bottom: best sequence by set-RNN. Above models, at each time, we list the top unigrams selected by attention.\vspace{2ex}}
\label{fig:models}
\end{figure}

Next we demonstrate the issue with prescribing the sequence order in seq2seq-RNN with a TheGuardian example\footnote{\scriptsize This document can be viewed at \url{http://www.guardian.co.uk/artanddesign/jonathanjonesblog/2009/apr/08/altermodernism-nicolas-bourriaud}}. Figure~\ref{fig:models} shows the predictions made by seq2seq-RNN and our method. In this particular example the top sequence agrees with the top set in our method's prediction so we can just analyze the top sequence. seq2seq-RNN predicts \texttt{Tate Modern} (incorrect but more popular label) while we predict \texttt{Tate Britain} (correct but less popular label). The seq2seq predicted sequence is in the decreasing label frequency order while our predicted sequence is not. In the training data, \texttt{Exhibition} is more frequent than \texttt{Tate Britain} and \texttt{Tate Modern}. If we arrange labels by decreasing frequency, \texttt{Exhibition} is immediately followed by \texttt{Tate Modern} 19 times, and by \texttt{Tate Britain} only 3 times. So it is far more likely to have \texttt{Tate Modern} than \texttt{Tate Britain} after \texttt{Exhibition}. However, at the set level, \texttt{Exhibition} and \texttt{Tate Modern} co-occurs 22 times while \texttt{Exhibition} and \texttt{Tate Britain} co-occurs 12 times, so the difference is not so dramatic. In this case, imposing the sequence order biases the probability estimation and leads to incorrect predictions.
\section{Conclusion}
In this work, we present an adaptation of RNN sequence models to the problem of multi-label classification for text. RNN only directly defines probabilities for sequences, but not for sets. Different from previous approaches, which either transform a set to a sequence in some pre-specified order, or relate the sequence probability to the set probability in some ad hoc way, our formulation is derived from a principled notion of set probability. We define the set probability as the sum of all corresponding sequence permutation probabilities. We derive a new training objective that maximizes the set probability and a new prediction objective that finds the most probable set. These new objectives are theoretically more appealing than existing ones, because they give the RNN model more freedom to automatically discover and utilize the best label orders.

\label{sec:conclusion}

\section*{Acknowledgements}
We thank reviewers and Krzysztof Dembczyński for their helpful comments,
Xiaofeng Yang for her help on writing, and Bingyu Wang for his help on proofreading. This work has been generously supported through a grant from the Massachusetts General Physicians Organization.

\bibliography{local,naaclhlt2019}

\begin{thebibliography}{25}
\expandafter\ifx\csname natexlab\endcsname\relax\def\natexlab#1{#1}\fi

\bibitem[{Abadi et~al.(2016)Abadi, Barham, Chen, Chen, Davis, Dean, Devin,
  Ghemawat, Irving, Isard, Kudlur, Levenberg, Monga, Moore, Murray, Steiner,
  Tucker, Vasudevan, Warden, Wicke, Yu, and
  Zheng}]{DBLP:conf/osdi/AbadiBCCDDDGIIK16}
Mart{\'{\i}}n Abadi, Paul Barham, Jianmin Chen, Zhifeng Chen, Andy Davis,
  Jeffrey Dean, Matthieu Devin, Sanjay Ghemawat, Geoffrey Irving, Michael
  Isard, Manjunath Kudlur, Josh Levenberg, Rajat Monga, Sherry Moore,
  Derek~Gordon Murray, Benoit Steiner, Paul~A. Tucker, Vijay Vasudevan, Pete
  Warden, Martin Wicke, Yuan Yu, and Xiaoqiang Zheng. 2016.
\newblock Tensorflow: {A} system for large-scale machine learning.
\newblock In \emph{12th {USENIX} Symposium on Operating Systems Design and
  Implementation, {OSDI} 2016, Savannah, GA, USA, November 2-4, 2016.}, pages
  265--283.

\bibitem[{Bahdanau et~al.(2014)Bahdanau, Cho, and
  Bengio}]{DBLP:journals/corr/BahdanauCB14}
Dzmitry Bahdanau, Kyunghyun Cho, and Yoshua Bengio. 2014.
\newblock Neural machine translation by jointly learning to align and
  translate.
\newblock \emph{CoRR}, abs/1409.0473.

\bibitem[{Boutell et~al.(2004)Boutell, Luo, Shen, and
  Brown}]{DBLP:journals/pr/BoutellLSB04}
Matthew~R. Boutell, Jiebo Luo, Xipeng Shen, and Christopher~M. Brown. 2004.
\newblock Learning multi-label scene classification.
\newblock \emph{Pattern Recognition}, 37(9):1757--1771.

\bibitem[{Chen et~al.(2018)Chen, Chen, Yeh, and
  Wang}]{DBLP:conf/aaai/ChenCYW18}
Shang{-}Fu Chen, Yi{-}Chen Chen, Chih{-}Kuan Yeh, and Yu{-}Chiang~Frank Wang.
  2018.
\newblock Order-free {RNN} with visual attention for multi-label
  classification.
\newblock In \emph{Proceedings of the Thirty-Second {AAAI} Conference on
  Artificial Intelligence, New Orleans, Louisiana, USA, February 2-7, 2018}.

\bibitem[{Cho et~al.(2014)Cho, van Merrienboer, G{\"{u}}l{\c{c}}ehre, Bahdanau,
  Bougares, Schwenk, and Bengio}]{DBLP:conf/emnlp/ChoMGBBSB14}
Kyunghyun Cho, Bart van Merrienboer, {\c{C}}aglar G{\"{u}}l{\c{c}}ehre, Dzmitry
  Bahdanau, Fethi Bougares, Holger Schwenk, and Yoshua Bengio. 2014.
\newblock Learning phrase representations using {RNN} encoder-decoder for
  statistical machine translation.
\newblock In \emph{Proceedings of the 2014 Conference on Empirical Methods in
  Natural Language Processing, {EMNLP} 2014, October 25-29, 2014, Doha, Qatar,
  {A} meeting of SIGDAT, a Special Interest Group of the {ACL}}, pages
  1724--1734.

\bibitem[{Dembczynski et~al.(2010)Dembczynski, Cheng, and
  H{\"{u}}llermeier}]{DBLP:conf/icml/DembczynskiCH10}
Krzysztof Dembczynski, Weiwei Cheng, and Eyke H{\"{u}}llermeier. 2010.
\newblock Bayes optimal multilabel classification via probabilistic classifier
  chains.
\newblock In \emph{Proceedings of the 27th International Conference on Machine
  Learning (ICML-10), June 21-24, 2010, Haifa, Israel}, pages 279--286.

\bibitem[{Ghamrawi and McCallum(2005)}]{ghamrawi2005collective}
Nadia Ghamrawi and Andrew McCallum. 2005.
\newblock Collective multi-label classification.
\newblock In \emph{Proceedings of the 14th ACM international conference on
  Information and knowledge management}, pages 195--200. ACM.

\bibitem[{Jin and Nakayama(2016)}]{DBLP:conf/icpr/JinN16}
Jiren Jin and Hideki Nakayama. 2016.
\newblock Annotation order matters: Recurrent image annotator for arbitrary
  length image tagging.
\newblock In \emph{23rd International Conference on Pattern Recognition, {ICPR}
  2016, Canc{\'{u}}n, Mexico, December 4-8, 2016}, pages 2452--2457.

\bibitem[{Kingma and Ba(2014)}]{DBLP:journals/corr/KingmaB14}
Diederik~P. Kingma and Jimmy Ba. 2014.
\newblock Adam: {A} method for stochastic optimization.
\newblock \emph{CoRR}, abs/1412.6980.

\bibitem[{Koyejo et~al.(2015)Koyejo, Natarajan, Ravikumar, and
  Dhillon}]{koyejo2015consistent}
Oluwasanmi~O Koyejo, Nagarajan Natarajan, Pradeep~K Ravikumar, and Inderjit~S
  Dhillon. 2015.
\newblock Consistent multilabel classification.
\newblock In \emph{Advances in Neural Information Processing Systems}, pages
  3321--3329.

\bibitem[{Li et~al.(2016)Li, Wang, Pavlu, and Aslam}]{li2016conditional}
Cheng Li, Bingyu Wang, Virgil Pavlu, and Javed Aslam. 2016.
\newblock Conditional bernoulli mixtures for multi-label classification.
\newblock In \emph{International Conference on Machine Learning}, pages
  2482--2491.

\bibitem[{Liu and Tsang(2015)}]{liu2015optimality}
Weiwei Liu and Ivor Tsang. 2015.
\newblock On the optimality of classifier chain for multi-label classification.
\newblock In \emph{Advances in Neural Information Processing Systems}, pages
  712--720.

\bibitem[{Lowerre(1976)}]{lowerre1976harpy}
Bruce~T Lowerre. 1976.
\newblock The harpy speech recognition system.
\newblock Technical report, CARNEGIE-MELLON UNIV PITTSBURGH PA DEPT OF COMPUTER
  SCIENCE.

\bibitem[{Mikolov et~al.(2013)Mikolov, Chen, Corrado, and
  Dean}]{DBLP:journals/corr/abs-1301-3781}
Tomas Mikolov, Kai Chen, Greg Corrado, and Jeffrey Dean. 2013.
\newblock Efficient estimation of word representations in vector space.
\newblock \emph{CoRR}, abs/1301.3781.

\bibitem[{Nam et~al.(2017)Nam, Loza~Menc\'{i}a, Kim, and
  F{\"{u}}rnkranz}]{DBLP:conf/nips/NamMKF17}
Jinseok Nam, Eneldo Loza~Menc\'{i}a, Hyunwoo~J. Kim, and Johannes
  F{\"{u}}rnkranz. 2017.
\newblock Maximizing subset accuracy with recurrent neural networks in
  multi-label classification.
\newblock In \emph{Advances in Neural Information Processing Systems 30: Annual
  Conference on Neural Information Processing Systems 2017, 4-9 December 2017,
  Long Beach, CA, {USA}}, pages 5419--5429.

\bibitem[{Read et~al.(2009)Read, Pfahringer, Holmes, and
  Frank}]{read2009classifier}
Jesse Read, Bernhard Pfahringer, Geoff Holmes, and Eibe Frank. 2009.
\newblock Classifier chains for multi-label classification.
\newblock In \emph{Joint European Conference on Machine Learning and Knowledge
  Discovery in Databases}, pages 254--269. Springer.

\bibitem[{Rumelhart et~al.(1988)Rumelhart, Hinton, Williams
  et~al.}]{rumelhart1988learning}
David~E Rumelhart, Geoffrey~E Hinton, Ronald~J Williams, et~al. 1988.
\newblock Learning representations by back-propagating errors.
\newblock \emph{Cognitive modeling}, 5(3):1.

\bibitem[{Tsoumakas and Katakis(2007)}]{tsoumakas2007multi}
Grigorios Tsoumakas and Ioannis Katakis. 2007.
\newblock Multi-label classification: An overview.
\newblock \emph{International Journal of Data Warehousing and Mining (IJDWM)},
  3(3):1--13.

\bibitem[{Vinyals et~al.(2016)Vinyals, Bengio, and Kudlur}]{vinyals2015order}
Oriol Vinyals, Samy Bengio, and Manjunath Kudlur. 2016.
\newblock Order matters: Sequence to sequence for sets.
\newblock \emph{CoRR}, abs/1511.06391.

\bibitem[{Wang et~al.(2018)Wang, Li, Pavlu, and Aslam}]{wang2018pipeline}
Bingyu Wang, Cheng Li, Virgil Pavlu, and Jay Aslam. 2018.
\newblock A pipeline for optimizing f1-measure in multi-label text
  classification.
\newblock In \emph{2018 17th IEEE International Conference on Machine Learning
  and Applications (ICMLA)}, pages 913--918. IEEE.

\bibitem[{Wang et~al.(2016)Wang, Yang, Mao, Huang, Huang, and
  Xu}]{DBLP:conf/cvpr/WangYMHHX16}
Jiang Wang, Yi~Yang, Junhua Mao, Zhiheng Huang, Chang Huang, and Wei Xu. 2016.
\newblock {CNN-RNN:} {A} unified framework for multi-label image
  classification.
\newblock In \emph{2016 {IEEE} Conference on Computer Vision and Pattern
  Recognition, {CVPR} 2016, Las Vegas, NV, USA, June 27-30, 2016}, pages
  2285--2294.

\bibitem[{Wang et~al.(2017{\natexlab{a}})Wang, Zhu, Gong, and
  Li}]{DBLP:journals/corr/abs-1709-08553}
Jingya Wang, Xiatian Zhu, Shaogang Gong, and Wei Li. 2017{\natexlab{a}}.
\newblock Attribute recognition by joint recurrent learning of context and
  correlation.
\newblock \emph{CoRR}, abs/1709.08553.

\bibitem[{Wang et~al.(2017{\natexlab{b}})Wang, Chen, Li, Xu, and
  Lin}]{DBLP:conf/iccv/WangCLXL17}
Zhouxia Wang, Tianshui Chen, Guanbin Li, Ruijia Xu, and Liang Lin.
  2017{\natexlab{b}}.
\newblock Multi-label image recognition by recurrently discovering attentional
  regions.
\newblock In \emph{{IEEE} International Conference on Computer Vision, {ICCV}
  2017, Venice, Italy, October 22-29, 2017}, pages 464--472.

\bibitem[{Yang et~al.(2018)Yang, Sun, Li, Ma, Wu, and
  Wang}]{DBLP:journals/corr/abs-1806-04822}
Pengcheng Yang, Xu~Sun, Wei Li, Shuming Ma, Wei Wu, and Houfeng Wang. 2018.
\newblock {SGM:} sequence generation model for multi-label classification.
\newblock \emph{CoRR}, abs/1806.04822.

\bibitem[{Zhang et~al.(2016)Zhang, Wu, Shen, Zhang, and
  Lu}]{DBLP:journals/corr/ZhangWSZL16}
Junjie Zhang, Qi~Wu, Chunhua Shen, Jian Zhang, and Jianfeng Lu. 2016.
\newblock Multi-label image classification with regional latent semantic
  dependencies.
\newblock \emph{CoRR}, abs/1612.01082.

\end{thebibliography}
\bibliographystyle{acl_natbib}

\appendix

\end{document}